\documentclass[journal,onecolumn,11pt]{IEEEtran} 
\ifCLASSINFOpdf
\else
\fi

\usepackage{graphicx}
\usepackage{sidecap}
\usepackage{xcolor}
\usepackage[nonumberlist,nogroupskip]{glossaries}
\usepackage{glossary-mcols}
\usepackage{comment}
\usepackage{wrapfig}
\usepackage[margin=1.25in]{geometry}
\usepackage{setspace}
\usepackage[numbers, square, comma, sort&compress]{natbib}
\usepackage[font=small]{caption}
\usepackage{mdframed}
\usepackage{subfig}
\DeclareCaptionType[fileext=box]{infobox}[Box]

\usepackage[utf8]{inputenc}

\newcommand{\reffig}[1]{{\color{blue!70}(Fig.~\ref{#1})}}

\newcommand{\refbox}[1]{{\color{blue!70}(Box.~\ref{#1})}}

\newcommand{\refsec}[1]{{Sec.~\ref{#1}}}

\newcommand{\Dp}[2][]{\frac{\partial #1}{\partial #2}}

\makeglossary
\renewcommand{\glossarysection}[2][]{}
\newacronym{sg}{SG}{surrogate gradient}
\newacronym{ann}{ANN}{artificial neural network}
\newacronym{rnn}{RNN}{recurrent neural network}
\newacronym{rcnn}{RCNN}{recurrently connected neural network}
\newacronym{snn}{SNN}{spiking neural network}
\newacronym{mlp}{MLP}{multi-layer perceptron}
\newacronym{stdp}{STDP}{spike timing-dependent plasticity}
\newacronym{bp}{BP}{backpropagation}
\newacronym{bptt}{BPTT}{backpropagation through time}
\newacronym{rtrl}{RTRL}{real-time recurrent learning}
\newacronym{lel}{LEL}{local error learning}
\newacronym{lif}{LIF}{leaky integrate-and-fire}
\newacronym{relu}{ReLU}{rectifying linear unit}
\newacronym{eRBP}{eRBP}{event-driven random backpropagation}
\newacronym{lstm}{LSTM}{long short-term memory}
\newacronym{psp}{PSP}{postsynaptic potential}
\newacronym{fa}{FA}{feedback alignment}
\newacronym{dfa}{dFA}{direct feedback alignment}

\begin{document}

\title{Surrogate Gradient Learning in Spiking Neural Networks}

\author{Emre O. Neftci$^\dagger$,~\IEEEmembership{Member,~IEEE,}
        Hesham Mostafa$^\dagger$,~\IEEEmembership{Member,~IEEE,}
        Friedemann Zenke$^\dagger$\\
        {\small $^\dagger$ All authors contributed equally. The order of authors is arbitrary.}}
{}
\maketitle
\IEEEpeerreviewmaketitle
%
\begin{abstract}
Spiking neural networks are nature's versatile solution to fault-tolerant and
energy efficient signal processing.
To translate these benefits into hardware, a growing number of neuromorphic
spiking neural network processors attempt to emulate biological neural networks. 
These developments have created an imminent need for methods and tools to
enable such systems to solve real-world signal processing problems. Like
conventional neural networks, spiking neural networks can be trained on real, domain specific data.  However, their
training requires overcoming a number of challenges linked to their binary
and dynamical nature.  This article elucidates step-by-step the problems
typically encountered when training spiking neural networks, and guides the
reader through the key concepts of synaptic plasticity and data-driven
learning in the spiking setting.  To that end, it gives an overview of
existing approaches and provides an introduction to surrogate gradient
methods, specifically, as a particularly flexible and efficient method to
overcome the aforementioned challenges. 
\end{abstract}

\section{Introduction}
Biological \glspl{snn} are evolution's highly efficient solution to the problem of signal processing.
Therefore, taking inspiration from the brain is a natural approach to engineering more efficient computing architectures.
In the area of machine learning, \glspl{rnn}, a class of stateful neural
networks whose internal state evolves with time \refbox{box:rnn}, 
have proven highly effective at solving real-time pattern recognition and noisy time series prediction problems \cite{Goodfellow_etal16_deeplear}.  
\Glspl{rnn} and biological neural networks share
several properties, such as a similar general architecture, temporal dynamics and learning
through weight adjustments.
Based on these similarities, a growing body of work is now establishing formal
equivalences between \glspl{rnn} and networks of spiking \gls{lif} neurons which are widely
used in computational neuroscience
\cite{zenke_superspike:_2018,bellec_long_2018,Kaiser_etal18_synaplas,tavanaei_deep_2018}. 
 
\Glspl{rnn} are typically trained using an optimization
procedure in which the parameters or weights are adjusted to minimize a given objective function.
Efficiently training large-scale \glspl{rnn} is challenging due to a variety of
extrinsic factors, such as noise and non-stationarity of the data, but also due to
the inherent difficulties of optimizing functions with long-range temporal and
spatial dependencies. 
In \glspl{snn} and binary \glspl{rnn}, these difficulties are compounded by the non-differentiable dynamics implied by the binary nature of their outputs.
While a considerable body of work has successfully 
demonstrated training of two-layer \glspl{snn}
\citep{gutig_spike_2014, memmesheimer_learning_2014,Anwani_Rajendran15_normappr}
without hidden units, and networks with recurrent synaptic connections
\citep{gilra_predicting_2017, nicola_supervised_2017},
the ability to train deeper \glspl{snn} with hidden layers has remained a major obstacle.
Because hidden units and depth are crucial to efficiently solve many real-world problems, overcoming this obstacle is vital.  

As network models grow larger and make their way into embedded and automotive applications, their power efficiency becomes increasingly important.
Simplified neural network architectures that can run natively and efficiently on dedicated hardware are now being devised.
This includes, for instance, networks of binary neurons or neuromorphic hardware that emulate the dynamics of \glspl{snn} \cite{boahen_neuromorphs_2017}.
Both types of networks dispense with energetically costly floating-point multiplications, making them particularly advantageous for low-power applications compared to neural networks executed on conventional hardware.

These new hardware developments have created an imminent need for tools and
strategies enabling efficient inference and learning in \glspl{snn} and binary \glspl{rnn}. 
In this article, we discuss and address the inherent difficulties in training
\glspl{snn} with hidden layers, and introduce various strategies and approximations used to
successfully implement them.

\section{Understanding \glsplural{snn} as \glsplural{rnn}}
\label{sec:understanding_ssn_as_rnn}

We start by formally mapping \glspl{snn}
to \glspl{rnn}. 
Formulating \glspl{snn} as \glspl{rnn} will allow us to directly transfer and
apply existing training methods for \glspl{rnn} and will serve as the conceptual
framework for the rest of this article. 

Before we proceed, one word on terminology.
We use the term \glspl{rnn} in its widest sense to refer to networks whose state evolves in time
according to a set of recurrent dynamical equations. Such dynamical recurrence 
can be due to the explicit presence of recurrent synaptic connections
between neurons in the network. This is the common understanding of
what a \gls{rnn} is.
But importantly, dynamical recurrence can also arise in the
\emph{absence} of recurrent connections. 
This happens, for instance, when stateful neuron or synapse models are used which
have internal dynamics. Because the network's state at a particular time step
recurrently depends on its state in previous time steps, these dynamics are
intrinsically recurrent. 
In this article, we use the term \gls{rnn} for networks exhibiting either,
or both types of recurrence. 
Moreover, we introduce the term \gls{rcnn} 
for the subset of networks with explicit recurrent synaptic connections.
We will now show that both admit the same mathematical treatment despite the
fact that their dynamical properties may be vastly different. 

To this end, we will first introduce the \gls{lif} neuron model with
current-based synapses which has wide use in computational neuroscience \cite{Gerstner_etal14_neurdyna}. 
Next, we will reformulate this model in discrete time and show its formal
equivalence to a \gls{rnn} with binary activation functions. 
Readers familiar with the \gls{lif} neuron model can skip the following steps up to Equation~\eqref{eq:mem_discrete_time}.

\begin{infobox}[tbhp]
  \begin{mdframed}[backgroundcolor=black!10]\small
	  \caption{\Glsfirstplural{rnn} \label{box:rnn}}
  \begin{wrapfigure}{r}{4em}
    \includegraphics[width=3.7em, trim={1.3cm 0 0 0}, clip]{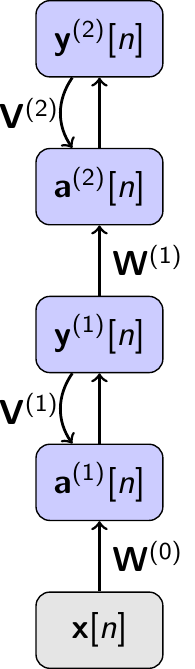}
  \end{wrapfigure}
	  \Glspl{rnn} are networks of inter-connected units, or neurons in which
	  the network state at any point in time ${a}[n]$ is a function of
	  both external input ${x}[n]$  and the network's state at the previous
	  time point  ${a}[n-1]$.
    One popular \gls{rnn} structure arranges neurons in multiple layers where every layer is recurrently connected and also receives input from the previous layer. More precisely, the dynamics of a network with $L$ layers is given by:
  \begin{align*}
	  {\bf y}^{(l)}[n] = & \sigma({\bf a}^{(l)}[n]) \quad \text{for $l=1,\ldots,L$}\\
	  {\bf a}^{(l)} [n] = &  {\bf V}^{(l)} {\bf y}^{(l)} [n-1]  +  {\bf W}^{(l)}
	  {\bf y}^{(l-1)} [n-1] \quad \text{for $l=1,\ldots,L$} \\
	  {\bf y}^{(0)}[n] \equiv & {\bf x}[n]
  \end{align*}
	  where ${\bf a}^{(l)}[n]$ is the state vector of the neurons at layer $l$,
	  $\sigma$ is an activation function, and $\mathbf{V}^{(l)}$ and $\mathbf{W}^{(l)}$ are
	  the recurrent and feedforward weight matrices of layer $l$, respectively.
	  External inputs ${\bf x}[n]$ typically arrive at the first layer. Non-scalar
	  quantities are typeset in bold face.
  \end{mdframed}
\end{infobox}

\begin{SCfigure}[10][tbhp]
    \centering
	\includegraphics[scale=0.9]{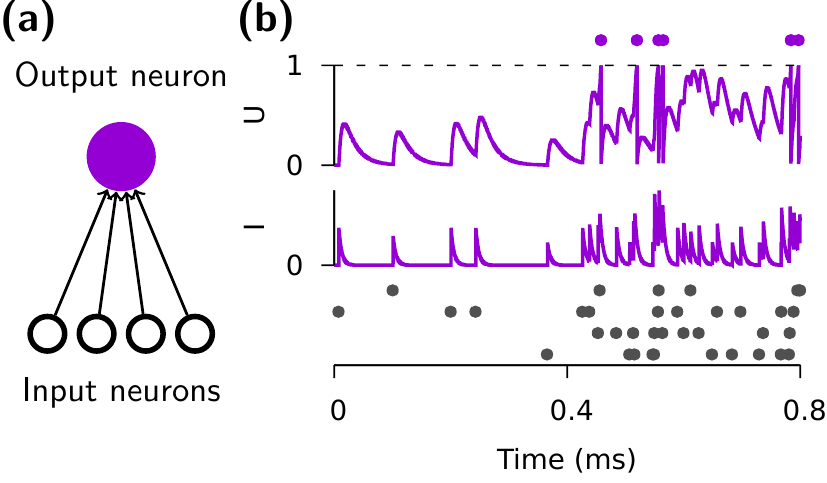}
	\caption{\textbf{Example \gls{lif} neuron dynamics.}
	(\textbf{a})~Schematic of network setup. Four input neurons connect
	to one postsynaptic neuron.
	(\textbf{b})~Input and output activity over time.
	Bottom panel: Raster plot showing the activity of the four input neurons.
	Middle panel: The synaptic current $I$.
	Top panel: The membrane potential $U$ of the output neuron as a function of
	time. 
	Output spikes are shown as points at the top.
	During the first 0.4\,s the dynamics are strictly ``sub-threshold'' and
	individual \glspl{psp} are clearly discernible.
	Only when multiple \glspl{psp} start to sum up, the neuronal firing
	threshold (dashed) is
	reached and output spikes are generated.
	}
    \label{fig:lif_neuron_activity}
\end{SCfigure}

A \gls{lif} neuron in layer $l$ with index $i$ can formally be described in differential form as
\begin{equation}
	\tau_\mathrm{mem} \frac{\mathrm{d}U_i^{(l)}}{\mathrm{d}t} = -(U_i^{(l)}-U_\mathrm{rest}) + RI_i^{(l)}
    \label{eq:lif_basic}
\end{equation}
where $U_i(t)$ is the membrane potential, $U_\mathrm{rest}$ is the resting potential, $\tau_\mathrm{mem}$ is the
membrane time constant, $R$ is the input resistance, and $I_i(t)$ is the 
input current \cite{Gerstner_etal14_neurdyna}.
Equation~\eqref{eq:lif_basic} shows that $U_i$ acts as a leaky integrator of the input current $I_i$.
Neurons emit spikes to communicate their output to other neurons when their membrane voltage reaches the firing threshold $\vartheta$.
After each spike, the membrane voltage $U_i$ is reset to the resting potential $U_\mathrm{rest}$ \reffig{fig:lif_neuron_activity}.
Due to this reset, Equation~\eqref{eq:lif_basic} only describes the subthreshold dynamics of a \gls{lif} neuron, \emph{i.e.}\ the dynamics in absence of spiking output of the neuron.

In \glspl{snn}, the input current is typically generated by synaptic currents
triggered by the arrival of presynaptic spikes $S_j^{(l)}(t)$.  
When working with differential equations, it is convenient to denote a spike
train $S_j^{(l)}(t)$ as a sum of Dirac delta functions $S_j^{(l)}(t)=\sum_{s \in
C_j^{(l)}} \delta(t-s)$ where $s$ runs over the firing times $C_j^{(l)}$ of
neuron $j$ in layer $l$.

Synaptic currents follow specific temporal dynamics
themselves. A common first-order approximation is to model their time course as an
exponentially decaying current following each presynaptic spike. Moreover, we assume that
synaptic currents sum linearly. The dynamics of these operations are given by
\begin{equation} 
	\frac{\mathrm{d}I_i^{(l)}}{\mathrm{d}t}=
	-\underbrace{\frac{I_i^{(l)}(t)}{\tau_\mathrm{syn}}}_\text{exp. decay} +\underbrace{\sum_j
	W_{ij}^{(l)} S_j^{(l-1)}(t)}_\mathrm{feed-forward} +\underbrace{\sum_j V_{ij}^{(l)}
	S_j^{(l)}(t)}_\mathrm{recurrent}
    \label{eq:cuba_syn}
\end{equation}
where the sum runs over all presynaptic neurons $j$ and $W_{ij}^{(l)}$ are the
corresponding afferent weights from the layer below. 
Further, the $V_{ij}^{(l)}$ correspond to explicit recurrent connections within
each layer.  
Because of this property we can simulate a
single \gls{lif} neuron with two linear differential equations whose initial
conditions change instantaneously whenever a spike occurs.
Through this property, we can incorporate the reset term in Equation~\eqref{eq:lif_basic} through an
extra term that instantaneously decreases the membrane potential by the amount $(\vartheta - U_\mathrm{rest})$ whenever the neuron emits a spike:
\begin{equation} 
	\frac{\mathrm{d}U_i^{(l)}}{\mathrm{d}t} = -\frac{1}{\tau_\mathrm{mem}}\left(
	(U_i^{(l)}-U_\mathrm{rest}) + RI_i^{(l)} \right) + S_i^{(l)}(t)(U_\mathrm{rest}-\vartheta)
    \label{eq:lif}
\end{equation}

It is customary to approximate the solutions of Equations~\eqref{eq:cuba_syn}
and~\eqref{eq:lif} numerically in discrete time and to express the output spike
train~$S_i^{(l)}[n]$ of neuron~$i$ in layer~$l$ at time step~$n$ as a nonlinear function of the
membrane voltage $S_i^{(l)}[n]\equiv\Theta(U_i^{(l)}[n]-\vartheta)$ where $\Theta$ denotes
the Heaviside step function and $\vartheta$ corresponds to the firing threshold. 
Without loss of generality, we
set $U_\mathrm{rest}=0$, $R=1$, and $\vartheta=1$.
When using a small simulation time step $\Delta_t>0$,
Equation~\eqref{eq:cuba_syn} is well approximated by
\begin{equation}
  I_i^{(l)}[n+1] = \alpha I_i^{(l)}[n] 
	+ \sum_j W_{ij}^{(l)} S_j^{(l)}[n] +
	\sum_j V_{ij}^{(l)} S_j^{(l)}[n]
    \label{eq:syn_discrete_time}
\end{equation}
with the decay strength 
$\alpha \equiv \exp\left(-\frac{\Delta_t}{\tau_\mathrm{syn}} \right)$. 
Note that $0<\alpha<1$ for finite and positive $\tau_\mathrm{syn}$.
Moreover, $S_j^{(l)}[n] \in \{0,1\}$.
We use $n$ to denote the time step to emphasize the discrete dynamics.
    We can now express Equation~\eqref{eq:lif} as 
\begin{equation}
	U_i^{(l)}[n+1] = \beta U_i^{(l)}[n] + I_i^{(l)}[n] -S_i^{(l)}[n]
    \label{eq:mem_discrete_time}
\end{equation}
with $\beta \equiv \exp\left(-\frac{\Delta_t}{\tau_\mathrm{mem}}\right)$.

\begin{SCfigure}[10][tbhp]
    \centering
    \includegraphics[scale=1.0]{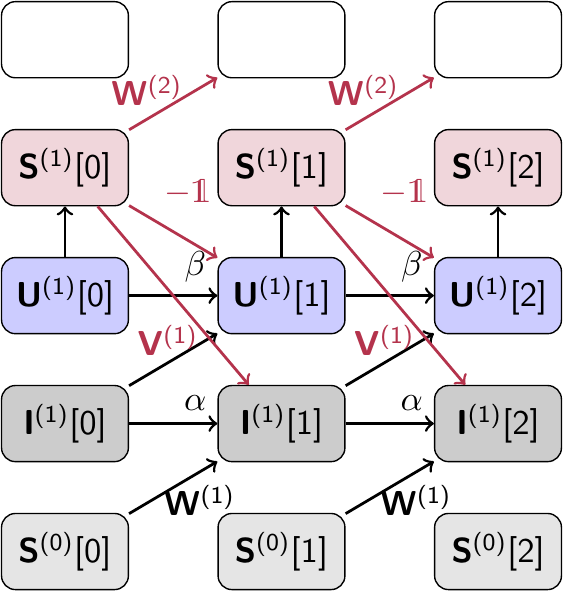}
	\caption{\textbf{Illustration of the computational graph of a \gls{snn}
	in discrete time.} Time steps flow from left to right. Input spikes
	$\mathbf{S}^{(0)}$ are fed
	into the network from the bottom and propagate upwards to higher layers. The
	synaptic currents $\mathbf{I}$ are decayed by $\alpha$ in each time step and fed
	into the membrane potentials $\mathbf{U}$. The $\mathbf{U}$ are similarly decaying over time
	as characterized by $\beta$. Spike trains $\mathbf{S}$ are generated by applying a
	threshold nonlinearity to the membrane potentials $\mathbf{U}$ in each time step.
	Spikes causally affect the network state (red connections).  First, each
	spike causes the membrane potential of the neuron that emits the spike to
	be reset. Second, each spike may be communicated to the same neuronal
	population via recurrent connections $\mathbf{V}^{(1)}$.  Finally, it may also be
	communicated via $\mathbf{W}^{(2)}$ to another downstream network layer
	or, alternatively, a readout layer on which a cost function is
	defined.}
    \label{fig:snn_computational_graph}
\end{SCfigure}

Equations~\eqref{eq:syn_discrete_time} and~\eqref{eq:mem_discrete_time}
characterize the dynamics of a \gls{rnn}. Specifically, the state of
neuron~$i$ is given by the instantaneous synaptic currents $I_i$ and the
membrane voltage $U_i$ \refbox{box:rnn}.  The computations necessary to update the cell state can
be unrolled in time as is best illustrated by the computational graph 
(Figure~\ref{fig:snn_computational_graph}).

We have now seen that \glspl{snn} constitute a special case of \glspl{rnn}.
However, so far we have not explained how their parameters are set to implement a specific computational function.
This is the focus of the rest of this article, in which we present a variety of
learning algorithms that systematically change the parameters towards
implementing specific functionalities. 

\section{Methods for training \glsplural{rnn}}
Powerful machine learning methods are able to train \glspl{rnn} for a variety of
tasks ranging from time series prediction, to language translation, to automatic
speech recognition \citep{Goodfellow_etal16_deeplear}.
In the following, we discuss the most common methods before analyzing their applicability to \glspl{snn}.

There are several stereotypical ingredients that define the training process.
The first ingredient is a cost or loss function which is minimized when the network's response corresponds to the desired behavior.
In time series prediction, for example, this loss could be the squared difference between the predicted and the true value.
The second ingredient is a mechanism that updates the network's weights to minimize the loss.
One of the simplest and most powerful mechanisms to achieve this is to perform
gradient descent on the loss function.
In network architectures with \emph{hidden units} (\emph{i.e.} units whose
activity affect the loss indirectly through other units) the parameter updates
contain terms relating to the activity and weights of the downstream units they project to. 
Gradient-descent learning solves this \emph{credit assignment problem} by providing explicit expressions for these updates through the chain rule of derivatives. 
As we will now see, the learning of hidden unit parameters depends on an efficient method to compute these gradients.
When discussing these methods, we distinguish between solving the spatial credit assignment problem which affects
\glspl{mlp} and \glspl{rnn} in the same way and the temporal credit assignment problem which only occurs in \glspl{rnn}.
We now discuss common algorithms which provide both types of credit assignment.

\subsection{Spatial credit assignment}
\label{sec:spatial_CA}
To train \glspl{mlp}, credit or blame needs to be assigned spatially across
layers and their respective units. 
This spatial credit assignment problem is solved most commonly by the
\gls{bp} of error algorithm \refbox{box:bp}. 
In its simplest form, this algorithm propagates errors ``backwards'' from the output of the network to upstream neurons.
Using \gls{bp} to adjust hidden layer weights ensures that the weight
update  will reduce the cost function for the current training example, provided the learning rate is small enough.
While this theoretical guarantee is desirable, it comes at the cost of certain
communication requirements --- namely that gradients have to be communicated
back through the network --- and increased memory requirements as the neuron
states need to be kept in memory until the errors become available.
\begin{infobox}[tbhp]
  \begin{mdframed}[backgroundcolor=black!10]\small
  \caption{\label{box:bp}\label{box:bptt} The Gradient Backpropagation Rule for Neural Networks} 

    The task of learning is to minimize a cost function $\mathcal{L}$ over the entire dataset.
    In a neural network, this can be achieved by gradient descent, which modifies the network parameters $\mathbf{W}$ in the direction opposite to the gradient:
    \begin{equation*}
      \begin{split}
        W_{ij} \leftarrow W_{ij} - \eta \Delta W_{ij},  & \text{where } \Delta W_{ij} =
        \Dp[\mathcal{L}]{W_{ij}} = 
       \Dp[\mathcal{L}]{y_i} 
       \Dp[y_i]{ a_i }       
       \Dp[a_i]{W_{ij}}      
      \end{split}
    \end{equation*}
    with $a_i = \sum_j W_{ij} x_j$ the total input to the neuron, $y_i$ is the output of neuron $i$, and $\eta$ a small learning rate. 
    The first term is the error of neuron $i$ and the second term reflects the sensitivity of the neuron output to changes in the parameter.
	  In multilayer networks, gradient descent is expressed as the \gls{bp} of the errors starting from the prediction (output) layer to the inputs. 
    Using superscripts $l=0,...,L$ to denote the layer ($0$ is input, $L$ is output):
    \begin{equation}\label{eq:bp_deep}
		\frac{\mathrm{\partial}}{\mathrm{\partial} W^{(l)}_{ij}} \mathcal{L} =
		\delta_{i}^{(l)}  y^{(l-1)}_j,\text{ where }\delta_{i}^{(l)} = \sigma'\left(
		a_i^{(l)} \right) \sum_k \delta_{k}^{(l+1)} W_{ik}^{\top,(l)},
    \end{equation}
	  where $\sigma'$ is the derivative of the activation function, and $\delta_{i}^{(L)}=\Dp[\mathcal{L}]{y_i^{(L)}}$ is the error of
	  output neuron $i$ and $y_{i}^{(0)}=x_i$ and $\top$ indicates the transpose.\\
   \begin{wrapfigure}{r}{15em}
   \includegraphics[width=15em]{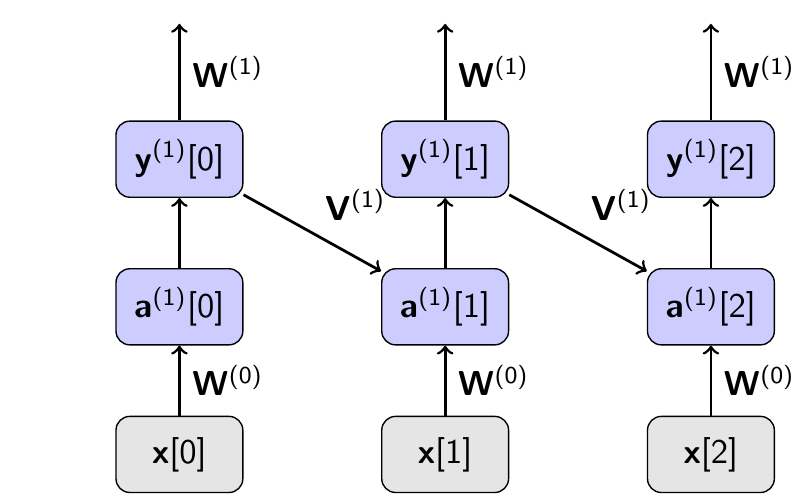}
   \caption*{``Unrolled'' RNN}
   \end{wrapfigure}  
    This update rule is ubiquitous in deep learning and known as the gradient
	  \gls{bp} algorithm \cite{Goodfellow_etal16_deeplear}.   
    Learning is typically carried out in forward passes (evaluation of the neural network activities) and backward passes (evaluation of $\delta$s).
    
    The same rule can be applied to \Glspl{rnn}. 
    In this case the recurrence is ``unrolled'' meaning that an auxiliary network is created by making copies of the network for each time step.
 
    The unrolled network is simply a deep network with shared feedforward
    weights $\mathbf{W}^{(l)}$ and recurrent weights $\mathbf{V}^{(l)}$, on which the standard \Gls{bp} applies:
    \begin{equation}\label{eq:bptt}
      \begin{split}
		  \Delta {W_{ij}^{(l)}} &\propto \frac{\mathrm{\partial}}{\mathrm{\partial}
		  W^{(l)}_{ij}} \mathcal{L}[n]  = \sum_{m=0}^t \delta_{i}^{(l)}[m]
		  y^{(l-1)}_j[m],\text{ and }\Delta {V_{ij}^{(l)}} \propto
		  \frac{\mathrm{\partial}}{\mathrm{\partial} V^{(l)}_{ij}}
		  \mathcal{L}[n]  = \sum_{m=1}^t \delta_{i}^{(l)}[s]  y^{(l)}_j[m-1]\\
		  \delta_{i}^{(l)} [n] & = \sigma'\left( a_i^{(l)}[n] \right) \left(
		  \sum_k \delta_{k}^{(l+1)}[n] W_{ik}^{\top,l} + \sum_k \delta_{k}^{(l)}[n+1] V_{ik}^{\top,l} \right),
      \end{split}
    \end{equation}
    Applying \gls{bp} to an unrolled network is referred to as \glsdesc{bptt} (BPTT).
  \end{mdframed}
\end{infobox}

\subsection{Temporal credit assignment} \label{sec:temporal_credit_assignment}
When training \glspl{rnn}, we also have to consider temporal interdependencies of
network activity. 
This requires solving the temporal credit assignment problem (Fig.~\ref{fig:snn_computational_graph}).
There are two common methods to achieve this:
\begin{enumerate}
\item The ``backward'' method:  
This method applies the same strategies as with spatial credit assignment by ``unrolling'' the network in	time \refbox{box:bptt}.
\Gls{bptt} solves the temporal credit assignment problem by back-propagating errors through the unrolled network.
This method works backward through time after completing a forward pass.
		The use of standard \gls{bp} on the unrolled network directly enables the
		use of autodifferentiation tools offered in modern machine learning
		toolkits~\cite{bellec_long_2018,shrestha_slayer:_2018}.

\item The forward method: In some situations, it is beneficial to propagate all necessary information for gradient computation forward in time \cite{Williams_Zipser89_learalgo}.
This formulation is achieved by computing the gradient of a cost function $\mathcal{L}[n]$ and maintaining the recursive structure of the \Gls{rnn}. For example, the ``forward gradient'' of the feed-forward weight $\mathbf{W}$ becomes:
\begin{equation} 
\begin{split} \label{eq:forward_mode_differentiation}
  \Delta {W_{ij}^m} &\propto \Dp[{\mathcal{L}[n]}]{W_{ij}^m}  = \sum_k
	\Dp[{\mathcal{L}[n]}]{y_k^{(L)}[n]} P_{ijk}^{L,m}[n],\text{ with }
	P_{ijk}^{(l,m)}[n]     = \frac{\partial} {\partial W_{ij}^m}  y_k^{(l)}[n]\\
	P_{ijk}^{(l,m)}[n] &= \sigma'(a^{(l)}_k[n]) \left( \sum_{j'} V_{ij'}^{(l)}
	P_{ijj'}^{(l,m)}[n-1] + \sum_{j'} W_{ij'}^{(l)} P_{ijj'}^{(l-1,m)}[n-1] +
	\delta_{lm} y_i^{(l-1)}[n-1] \right). \\
\end{split}
\end{equation}
\noindent Gradients with respect to recurrent weights $V_{ij}^{(l)}$ can be computed in a similar fashion \cite{Williams_Zipser89_learalgo}.
\end{enumerate}
The backward optimization method is generally more efficient in terms of computation, but requires maintaining all the inputs and activations for each time step.
Thus, its space complexity for each layer is $O(N T)$, where $N$ is the number
of neurons per layer and $T$ is the number of time steps.
On the other hand, the forward method requires maintaining variables
$P_{ijk}^{(l,m)}$, resulting in a $O(N^3)$ space complexity per layer. 
While $O(N^3)$ is not a favorable scaling compared to $O(NT)$ for large $N$,
simplifications of the computational graph can reduce the memory complexity of
the forward method to $O(N^2)$ \cite{bellec_biologically_2019,
zenke_superspike:_2018}, or even $O(N)$\cite{Kaiser_etal18_synaplas}. These
simplifications also reduce the computational complexity, rendering the scaling
of forward algorithms comparable or better than \gls{bptt}. 
Such simplifications are at the core of several successful approaches which we
will describe in \refsec{sec:applications}.
Furthermore, the forward method is more appealing from a biological point of
view, since the learning rule can be made consistent with synaptic plasticity in
the brain and ``three-factor'' rules, as discussed in Section \ref{sec:superspike}.

In summary, efficient algorithms to train \glspl{rnn} exist.
We will now focus on training \glspl{snn}.

\section{Credit assignment with spiking neurons: Challenges and solutions}
\begin{SCfigure}[10][tbhp]
  \centering
  \includegraphics[scale=1]{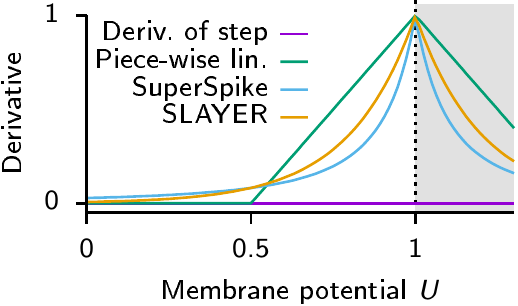}
  \caption{\textbf{Commonly used surrogate derivatives}. 
	The step function has zero derivative (violet) everywhere except at~0 where it is ill defined.
	Examples of surrogate derivatives  
	which have been used to train \glspl{snn}.
	Green: Piece-wise linear \citep{esser_convolutional_2016,bellec_long_2018,bohte_error-backpropagation_2011}.
	Blue: Derivative of a fast sigmoid \citep{zenke_superspike:_2018}.
	Yellow: Exponential \citep{shrestha_slayer:_2018}.
	Note that the axes have been rescaled on a per-function-basis for illustration
	purposes. 
	\label{fig:surr_partials}}
\end{SCfigure}

So far we have discussed common algorithmic solutions to training \glspl{rnn}.
Before these solutions can be applied to \glspl{snn}, however, two key challenges need to
be overcome. The first challenge concerns the non-differentiability of the spiking nonlinearity.
Equations~\eqref{eq:bptt} and~\eqref{eq:forward_mode_differentiation} reveal that the expressions for both the forward and the backward learning methods contain the 
derivative of the neural activation function $\sigma' \equiv
\Dp[y_i^{(l)}]{a_i^{(l)}}$ as a multiplicative factor.
For a spiking neuron, however, we have $S(U(t))=\Theta(U(t)-\vartheta)$, whose
derivative is zero everywhere except at $U=\vartheta$, where it is ill defined \reffig{fig:surr_partials}.

This all-or-nothing behavior of the binary spiking nonlinearity stops gradients
from ``flowing'' and makes \gls{lif} neurons unsuitable for gradient based optimization.
The same issue occurs in binary neurons and some of the solutions proposed here are inspired by the methods first developed in binary networks \cite{courbariaux_binarized_2016,bengio_estimating_2013}. 

The second challenge concerns the implementation of the optimization algorithm itself. 
Standard \gls{bp} can be expensive in terms of computation, memory and
communication, and may be poorly suited to the constraints dictated by the
hardware that implements it (e.g. a computer, a brain, or a neuromorphic device).
Processing in dedicated neuromorphic hardware and, more generally, non-von Neumann computers may have specific locality requirements \refbox{box:nonlocal} that can complicate matters. 
On such hardware, the forward approach may therefore be preferable.
In practice, however, the scaling of both methods ($O(N^3)$ and $O(NT)$) has proven unsuitable for many \gls{snn} models. 
For example, the size of the convolutional \gls{snn} models trained with \gls{bptt} for gesture classification \cite{Shrestha_Orchard18_slayspik} are GPU memory bounded.
Additional simplifying approximations that reduce the complexity of the forward method will be discussed below.
In the following sections, we describe approximate solutions to these challenges that make learning in \glspl{snn} more tractable.

To overcome the first challenge in training \glspl{snn}, which is 
concerned with the discontinuous spiking nonlinearity, several approaches have been devised with varying degrees of success.
The most common approaches can be coarsely classified into the following categories: 
  i)~resorting to entirely biologically inspired local learning rules for the hidden units, 
 ii)~translating conventionally trained ``rate-based'' neural networks to
 \glspl{snn},
iii)~smoothing the network model to be continuously differentiable, or 
 iv)~defining a \gls{sg} as a continuous relaxation of the real
gradients.  
Approaches pertaining biologically motivated local learning rules (i)
and network translation (ii) have been
reviewed extensively elsewhere \citep{abbott_building_2016,tavanaei_deep_2018}. 
In this article, we therefore focus on the latter two supervised approaches (iii
\& iv)
which we will refer to as the ``smoothed'' and the \gls{sg} approach.
First, we review existing literature on common ``smoothing'' approaches before turning to an in-depth discussion of how to build
functional \glspl{snn} using \gls{sg} methods.

\subsection{Smoothed spiking neural networks}
The defining characteristic of smoothed \glspl{snn} is that their formulation ensures well-behaved gradients which are directly suitable for optimization.
Smooth models can be further categorized into 
(1)~soft nonlinearity models, 
(2)~probabilistic models, for which gradients are only well defined in expectation, or models which either rely entirely on 
(3)~rate or
(4)~single-spike temporal codes.

\subsubsection{Gradients in soft nonlinearity models}

This approach can in principle be applied
directly to all spiking neuron models which explicitly include a smooth spike
generating process. This includes, for instance,  the Hodgkin-Huxley,
Morris-Lecar, and FitzHugh-Nagumo models \cite{Gerstner_etal14_neurdyna}.
In practice this approach has only been applied successfully by
\citet{huh_gradient_2018} using an augmented integrate-and-fire model in
which the binary spiking nonlinearity was replaced by a continuous-valued
gating function.  The resulting network constitutes a \gls{rcnn} which can be
optimized using standard methods of \gls{bptt} or \gls{rtrl}.  
Importantly, the soft threshold models compromise on one of
the key features of \gls{snn}, namely the binary spike propagation.

\subsubsection{Gradients in probabilistic models}
Another example for smooth models are binary probabilistic models. 
In simple terms, stochasticity effectively smooths out the discontinuous binary
nonlinearity which makes it possible to define a gradient on expectation values.
Binary probabilistic models have been objects of extensive study in the machine
learning literature mainly in the context of (restricted) Boltzmann machines \cite{ackley_learning_1985}. 
Similarly, the propagation of gradients has been studied for binary stochastic
models \cite{bengio_estimating_2013}. 

Probabilistic models are practically useful because the
log-likelihood of a spike train is a smooth quantity which can be optimized using gradient descent
\citep{pfister_optimal_2006}.
Although this insight was first discovered in networks without hidden units,
the same ideas were later extended to multi-layer networks \citep{gardner_learning_2015}.
Similarly, \citet{guerguiev_towards_2017} used probabilistic neurons to study 
biologically plausible ways of propagating error or target signals using
segregated dendrites (see Section~\ref{sec:feedback_alignment}). 
In a similar vein, variational learning approaches were shown to be capable
of learning useful hidden layer representations in \glspl{snn}
\cite{brea_matching_2013, rezende_stochastic_2014, Mostafa_Cauwenberghs18}.
However, the injected noise necessary to smooth out the effect
of binary nonlinearities often poses a challenge for optimization
\cite{rezende_stochastic_2014}.
How noise, which is found ubiquitously in neurobiology, 
influences learning in the brain, remains an open question.

\subsubsection{Gradients in rate-coding networks}

Another common approach to obtain gradients in \glspl{snn} is to
assume a rate-based coding scheme.
The main idea is that spike rate is the underlying information-carrying quantity.
For many plausible neuron models, the supra-threshold firing rate depends
smoothly on the neuron input. This input-output dependence is captured by the
so-called f-I curve of a neuron.
In such cases, the derivative of the f-I curves is suitable for gradient-based optimization.

There are several examples of this approach. 
For instance, \citet{Hunsberger_Eliasmith15_spikdeep} as well as \citet{Neftci_etal17_evenranda} used an effectively
rate-coded input scheme to demonstrate competitive performance on standard
machine learning benchmarks such as CIFAR10 and MNIST.
Similarly \citet{Lee_etal16_traideep} demonstrated deep learning in \glspl{snn} 
by defining partial derivatives on low-pass filtered spike
trains. 

Rate-based approaches can offer good performance, but they may be inefficient.
On the one hand, precise estimation of firing rates requires averaging over
a number of spikes.  Such averaging requires either relatively high firing rates or
long averaging times because several repeats are needed to average out discretization noise.  
This problem can be partially addressed by spatial averaging over
large populations of spiking neurons. However, this may require the use of
larger neuron numbers.

Finally, the distinction between rate-coding and probabilistic networks can be blurry
since many probabilistic network implementations use rate-coding at the
output level.
Both types of models are differentiable, but for different reasons:
Probabilistic models are based on a firing probability densities
\cite{pfister_optimal_2006}. 
Importantly, the firing probability of a neuron is a continuous function. 
Although measuring probability changes requires ``trial averaging'' over several samples, it is the underlying continuity of the probability
density which formally allows to define 
differential improvements and thus to derive gradients.
By exploiting this feature, probabilistic models have been used to learn precise
output spike timing \citep{pfister_optimal_2006, gardner_learning_2015}.
In contrast, deterministic networks always emit a fixed integer number of spikes
for a given input. To nevertheless get at a notion of differential improvement, one 
may consider the number of spikes over a given time interval within single
trials. When
averaging over sufficiently large intervals, the resulting firing rates
behave as a quasi continuous function of the input current. This smooth input
output relationship is captured by the neuronal f-I~curve which can be used for
optimization \citep{Hunsberger_Eliasmith15_spikdeep,Neftci_etal17_evenranda}. 
Operating at the level of rates, however, comes at the expense temporal precision.

\subsubsection{Gradients in single-spike-timing-coding networks}
In an effort to optimize \glspl{snn} without potentially harmful noise
injection and without reverting to a rate-based coding scheme, 
several studies have considered the outputs of neurons in \glspl{snn} to be a set of firing times.
In such a temporal coding setting, individual spikes could carry significantly more information than rate-based schemes that only consider the total number of spikes in an interval.
 
The idea behind training temporal coding networks was pioneered in SpikeProp \citep{bohte_error-backpropagation_2002}.
In this work the analytic expressions of firing times for hidden units were linearized, allowing to analytically compute approximate hidden layer gradients. 
More recently, a similar approach without the need for linearization 
was used in \cite{Mostafa16_supelear} where the author computed the spike timing 
gradients explicitly for non-leaky integrate-and-fire neurons. 
Intriguingly, the work showed competitive performance on conventional networks and benchmarks.

Although the spike timing formulation does in some cases
yield well-defined gradients, it may suffer from certain
limitations. 
For instance, the formulation of SpikeProp \citep{bohte_error-backpropagation_2002} required each hidden
unit to emit exactly one spike per trial, because it is impossible to define 
firing time for quiescent units. Ultimately, such a non-quiescence requirement could be
at conflict with power-efficiency for which it is conceivably beneficial to, for
instance, only have a subset of neurons active for any given task.

\subsection{Surrogate gradients}
\label{sec:surrogate_gradients}

\begin{figure}[tbhp]
  \centering
  \includegraphics[width=3.0in]{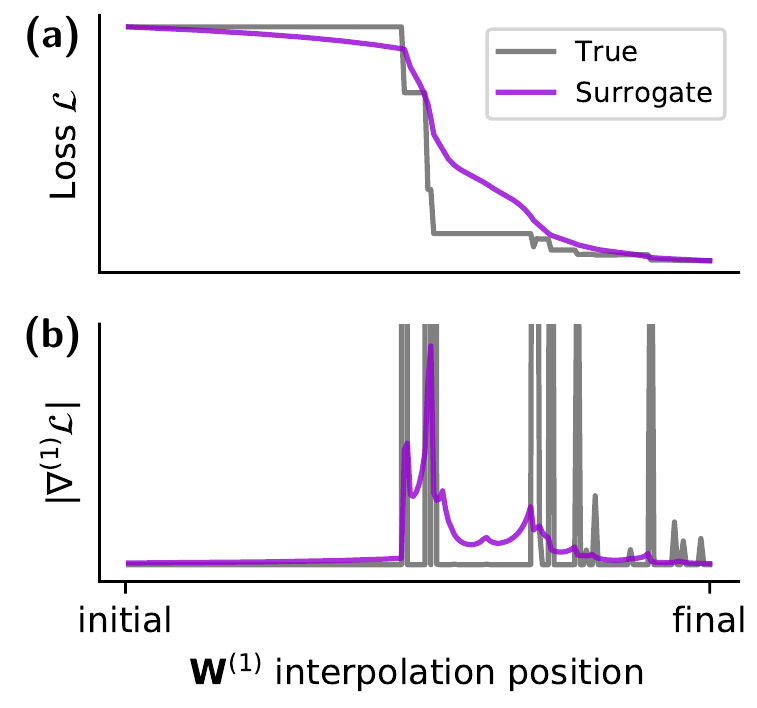}
	\caption{\textbf{Example of \gls{sg} for a \gls{snn} classifier.} 
	(\textbf{a})~Value of the loss function (gray) of an \gls{snn} classifier
	along an interpolation path over the hidden layer parameters $\mathbf{W}^{(1)}$.
	Specifically, we linearly interpolated between the random initial and
	final (post-optimization) weight matrices of the hidden layer inputs $\mathbf{W}^{(1)}$
	(network details: 2~input, 2~hidden, and 2~output units trained on a binary classification
	task).
	Note that the loss function (gray) displays characteristic plateaus with zero
	gradient which are detrimental for numerical optimization.
	(\textbf{b})~Norm of hidden layer (surrogate) gradients in arbitrary units
	along the interpolation path. To perform numerical optimization in
	this network we constructed
	a~\gls{sg} (violet) which, in contrast to the true gradient (gray), is non-zero.
	Note that we obtained the ``true gradient'' via the finite differences method which in itself is an approximation.
	Importantly, the \gls{sg} approximates the true gradient, but retains favorable properties for optimization, i.e.\ 
	continuity and finiteness.
	The \gls{sg} can be thought of as the gradient of a \textit{virtual} surrogate loss
	function (violet curve in (a); obtained by numerical integration of the
	\gls{sg} and scaled to match loss at initial and final point). This surrogate loss remains
	virtual because it is generally not computed explicitly. 
	In practice, suitable \glspl{sg} are obtained
	directly from the gradients of the original network through sensible
	approximations.  
	This is a key difference with respect to some other approaches \citep{huh_gradient_2018} in which the entire network is replaced explicitly by a surrogate network on which gradient descent can be performed using its true
	gradients.}
	\label{fig:surrgrad_concept}
\end{figure}

\gls{sg} methods provide an alternative approach to overcoming the difficulties
associated with the discontinuous nonlinearity. Moreover, they hold 
opportunities to reduce the potentially high algorithmic complexity associated
with training \glspl{snn}.
Their defining characteristic is that instead of changing the model definition
as in the smoothed approaches, a \gls{sg} is introduced.
In the following we make two distinctions.
We first consider \glspl{sg} which constitute a continuous relaxation of the non-smooth spiking
nonlinearity for purposes of numerical optimization \reffig{fig:surrgrad_concept}.
Such \glspl{sg} do not explicitly change the optimization algorithm itself and
can be used, for instance, in combination with \gls{bptt}.
Further, we also consider \glspl{sg} with more profound changes that explicitly affect
locality of the underlying optimization algorithms themselves to improve the computational and/or memory access overhead of the learning process.
One example of this approach that we will discuss involves replacing the global loss by a number of local loss functions.
Finally, the use of \glspl{sg} allows to efficiently train \glspl{snn} end-to-end without
the need to specify which coding scheme is to be used in the hidden 
layers. 

Like standard gradient-descent, \gls{sg} learning can deal with the spatial and
temporal credit assignment problem by either \gls{bptt} or by forward methods, e.g.\ through the use of eligibility
traces (see Section~\ref{sec:temporal_credit_assignment} for details).
Alternatively, additional approximations can be introduced which may offer advantages specifically for hardware implementations. 
In the following, we briefly review existing work relying on \gls{sg} methods before turning to a more in-depth treatment of the underlying principles and capabilities.

\subsubsection{Surrogate derivatives for spiking nonlinearity}
A set of works have used \gls{sg} to specifically overcome the challenge of the
discontinuous spiking nonlinearity. 
In these works, typically a standard algorithm such as \gls{bptt} is used with one minor modification: within the algorithm each occurrence of the derivative of the spiking nonlinearity is replaced by the derivative of a smooth function.
Implementing these approaches is straight-forward in most auto-differentiation-enabled machine learning toolkits.

One of the first uses of such a \gls{sg} is described in
\citet{bohte_error-backpropagation_2011} where the derivative of a
spiking neuron non-linearity was approximated by the derivative of a truncated quadratic
function, thus resulting in a \gls{relu} as surrogate derivative
\reffig{fig:surr_partials}.
This is similar in flavor to the solution proposed to optimize binary neural networks \citep{courbariaux_binarized_2016}. 
The same idea underlies the training of large-scale
convolutional networks with binary activations on classification problems using
neuromorphic hardware \citep{esser_convolutional_2016}.
\citet{zenke_superspike:_2018} proposed a three factor online learning rule using
a fast sigmoid to construct a \gls{sg}. \citet{shrestha_slayer:_2018} used an exponential function and
reported competitive performance on a range of neuromorphic benchmark problems.
Additionally, \citet{Oconnor_etal17} described a spike-based encoding method inspired
by Sigma-Delta modulators. They used their method to approximately encode both
the activations and the errors in standard feedforward \glspl{ann}, and apply standard backpropagation on these sparse approximate encodings.

Surrogate derivatives have also been used to train spiking \glspl{rcnn} where
dynamical recurrence arises due to the use of \gls{lif} neurons as well as due
to recurrent synaptic connections.  
Recently, \citet{bellec_long_2018} successfully trained \glspl{rcnn} with 
slow temporal neuronal dynamics using a piecewise linear surrogate derivative. 
Encouragingly, the  authors found that such networks can perform on par with
conventional \gls{lstm} networks.  
Similarly, \citet{Wozniak_etal18_deepnetw} reported competitive performance on a series of temporal benchmark datasets.

In summary, a plethora of studies have constructed \gls{sg} using different 
nonlinearities and trained a diversity of \gls{snn} architectures.
These nonlinearties, however, have a common underlying theme.
All functions are nonlinear and monotonically increasing towards the firing
threshold \reffig{fig:surr_partials}.
While a more systematic comparison of different surrogate nonlinearities is
still pending, overall the diversity found in the present literature suggests that the success of the method is not crucially dependent on the
details of the surrogate used to approximate the derivative.

\subsubsection{Surrogate gradients affecting locality of the update rules}

The majority of studies discussed in the previous section introduced a surrogate
nonlinearity to prevent gradients from vanishing (or exploding), but by relying on 
methods such as \gls{bptt}, they did not explicitly affect the structural properties of the learning rules. 
There are, however, training approaches for \glspl{snn} which introduce more
far-reaching modifications which may completely alter the way error signals
or target signals are propagated (or generated) within the network. 
Such approaches are typically used in conjunction with the aforementioned surrogate derivatives. 
There are two main motivations for such modifications which are typically linked
to physical constraints that make it impossible to implement the ``correct'' gradient descent algorithm. 
For instance, in neurobiology biophysical constraints make it impossible to implement \gls{bptt} without further approximations. 
Studies interested in how the brain could solve the credit assignment problem
focus on how simplified ``local'' algorithms could achieve similar performance
while adhering to the constraints of the underlying biological wetware \refbox{box:nonlocal}.
Similarly, neuromorphic hardware may pose certain constraints with regard to
memory or communications which impede the use of \gls{bptt} and call for simpler
and often more local methods for training on such devices.

As training \glspl{snn} using \glspl{sg} advances to deeper architectures, 
it is foreseeable that additional problems, similar to the ones
encountered in \glspl{ann}, will arise.
For instance, several approaches currently rely on \glspl{sg} derived
from sigmoidal activation functions (Fig.~\ref{fig:surr_partials}). 
However, the use of sigmoidal activation functions is implicated with vanishing
gradient problems.
Another set of challenges which may well need tackling in the future could be 
linked to the bias which \glspl{sg} introduce into the learning dynamics.

In the following Applications Section, we will review a selection of promising
\gls{sg} approaches which introduce far larger deviations from the ``true
gradients'' and still allow for learning at a greatly reduced complexity and
computational cost.

\begin{SCfigure}
  \centering
  \subfloat[BP]{\includegraphics[scale=.75]{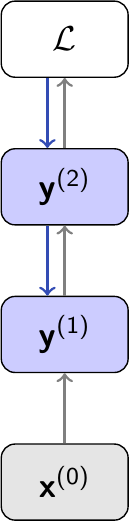}}
	\qquad
  \subfloat[FA]{\includegraphics[scale=.75]{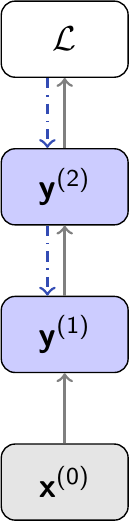}}
	\qquad
  \subfloat[DFA]{\includegraphics[scale=.75]{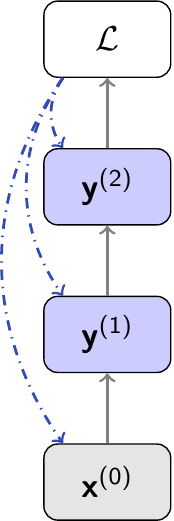}}
	\qquad
  \subfloat[Local Errors]{\includegraphics[scale=.75]{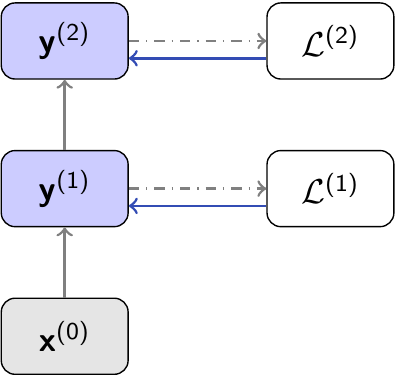}}
  \caption{\textbf{Strategies for relaxing gradient \Gls{bp} requirements}. Dashed lines indicate fixed, random connections. (a) \Gls{bp} propagates errors through each layer using the transpose of the forward weights by alternating forward and backward passes. (b) Feedback Alignment \cite{Lillicrap_etal16_randsyna} replaces the transposed matrix with a random one. (c) Direct Feedback Alignment \cite{nokland_direct_2016} directly propagates the errors from the top layer to the hidden layers. (d) Local errors \cite{Mostafa_Cauwenberghs18} uses a fixed, random, auxiliary cost function at each layer.}
  
  \label{fig:spatial_credit_assignment}
\end{SCfigure}

\section{Applications}\label{sec:applications}
In this section, we present a selection of illustrative applications of smooth or
\glspl{sg} to \glspl{snn} which exploit both the internal continuous-time dynamics of the neurons 
and their event-driven nature. 
The latter allows a network to remain quiescent until incoming spikes trigger activity. 

\subsection{Feedback alignment and random error \glsdesc{bp}}
\label{sec:feedback_alignment}

One family of algorithms that relaxes some of the requirements of \gls{bp} are
feedback alignment or, more generally, random \gls{bp} algorithms
\cite{Lillicrap_etal16_randsyna,Baldi_Sadowski16_theoloca,nokland_direct_2016}. 
These are approximations to the gradient \gls{bp} rule that side-step the
non-locality problem by replacing weights in the \gls{bp} rule with random ones
(Fig.~\ref{fig:spatial_credit_assignment}b):
$
\delta_{i}^{(l)} = \sigma'\left( a_i^{(l)} \right) \sum_k \delta_{k}^{(l+1)}
G_{ki}^{(l)},
$
where $\mathbf{G}^(l)$ is a fixed, random matrix with the same dimensions as ${\bf W}$. 
The replacement of $\mathbf{W}^{\top,(l)}$ with a random matrix $\mathbf{G}^{(l)}$ breaks the
dependency of the backward phase on $\mathbf{W}^{(l)}$, enabling the rule to be more local. 
One common variation is to replace the entire backward propagation by a random
propagation of the errors to each layer
(Fig.~\ref{fig:spatial_credit_assignment}c) \cite{nokland_direct_2016}:
$
\delta_{i}^{(l)} = \sigma'\left( a_i^{(l)} \right) \sum_k \delta^{(L)}_k
H_{ki}^{(l)},
$
where $\mathbf{H}^{(l)}$ is a fixed, random matrix with appropriate dimensions.

Random \Gls{bp} approaches lead to remarkably little loss in classification
performance on some benchmark tasks.
Although a general theoretical understanding of random \Gls{bp} is still a
subject of intense research, simulation studies have shown that, during learning, the network adjusts its feed-forward weights such that they partially align with the (random) feedback weights, thus permitting them to convey useful error information \cite{Lillicrap_etal16_randsyna}.
Building on these findings, an asynchronous spike-driven adaptation of random \gls{bp} using local synaptic plasticity rules with the dynamics of spiking neurons was demonstrated in \cite{Neftci_etal17_evenranda}.
To obtain the \glspl{sg}, the authors approximated the derivative of the neural activation function using a symmetric function that is zero everywhere except in the vicinity of zero, where it is constant. The derivative of this function exists and is piecewise constant.
Networks using this learning rule performed remarkably well, and were shown to operate continuously and asynchronously without the alternation between forward and backward passes that is necessary in \gls{bp}.
One important limitation with random \gls{bp} applied to \glspl{snn} was that the temporal dynamics of the neurons and synapses was not taken into account in the gradients. The following rule, SuperSpike solves this problem.

\begin{infobox}[tbhp]
  \begin{mdframed}[backgroundcolor=black!10]\small
  \caption{\label{box:nonlocal} Local models of computation}
  \begin{wrapfigure}{r}{10em}     \includegraphics[width=10.5em]{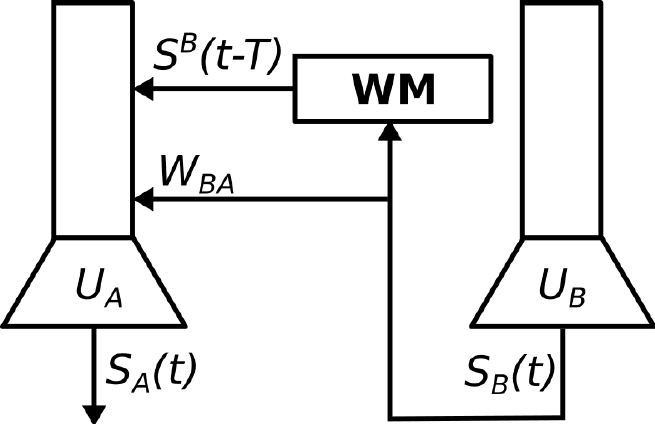}   \end{wrapfigure}
     Locality of computations is characterized by the set variables available to the physical processing elements, and depends on the computational substrate.
    To illustrate the concept of locality, we assume two neurons, $A$ and $B$, and would like Neuron $A$ to implement a function on domain $D$ defined as:
    \[ 
      \begin{split}
        D & = D_{loc} \cup D_{nloc},  \text{where } D_{loc}=\{W_{BA},S_A(t), U_A(t)\}\\\text{ and }D_{nloc} &= \{ S_B(t-T), U_{B}\}.
      \end{split}
    \]
    Here, $S^B(t-T)$ refers to the output of neuron $B$ $T$ seconds ago, $U_A$, $U_B$ are the respective membrane potentials, and $W_{BA}$ is the synaptic weight from $B$ to $A$.  
    Variables under $D_{loc}$ are directly available to Neuron A and are thus local to it. 
      
    On the other hand, variable $S^B(t-T)$ is temporally non-local and $U_{B}$ is spatially non-local to neuron $A$.
    Non-local information can be transmitted through special structures, for example dedicated encoders and decoders for $U_B$ and a form of working memory (WM) for $S_B(t-T)$.
    Although locality in a model of computation can make its use challenging, it enables massively parallel computations with dynamical interprocess communications. 
  \end{mdframed}
\end{infobox}

\subsection{Supervised learning with local three factor learning rules} 
\label{sec:superspike}

SuperSpike is a
biologically plausible three factor learning rule. 
In contrast to many existing three factor rules which fall into the category of
``smoothed approaches''
\citep{pfister_optimal_2006, gardner_learning_2015, guerguiev_towards_2017,
brea_matching_2013, rezende_stochastic_2014, Mostafa_Cauwenberghs18}, 
SuperSpike is a \gls{sg} approach which combines several 
approximations to render it more biologically plausible~\citep{zenke_superspike:_2018}.
Although the underlying motivation of the study is geared toward a deeper
understanding of learning in biological neural networks, the learning rule may
prove interesting for hardware implementations because it does not rely on \gls{bptt}.
Specifically, the rule uses synaptic eligibility traces to
solve the temporal credit assignment problem. 

We now provide a short account on why SuperSpike can be seen as one of
the forward-in-time optimization procedures. 
SuperSpike was derived for temporal supervised learning tasks in which a given
output neuron learns to spike at predefined times. To that end, SuperSpike
minimizes the van Rossum distance with kernel $\epsilon$ between a set of output
spike train $S_k(t)$ and their corresponding
target spike trains $S_k^*(t)$
\begin{equation}
	\mathcal{L} = \frac{1}{2} \int_{-\infty}^t \mathcal{L}(s)~ ds = \frac{1}{2}
	\int_{-\infty}^t \left( \epsilon\ast(S_k(s)-S_k^*(s)) \right)^2 ds \approx
	\frac{1}{2} \sum_n \left( \epsilon\ast(S_k[n]-S_k^*[n]) \right)^2
\end{equation}
where the last approximation corresponds to transitioning to discrete time. 
To perform online gradient descent, we need to compute the
gradients of $\mathcal{L}[n]$.
Here we first encounter the derivative
$\frac{\partial}{\partial W_{ij}} \epsilon \ast S_k[n]$.
Because the (discrete) convolution is a linear operator, this expression simplifies to 
$\epsilon \ast \frac{\partial S_k[n]}{\partial W_{ij}}$.
In SuperSpike $\epsilon$ is implemented as a dynamical system
(see \citep{zenke_superspike:_2018} for details).
To compute derivatives of the neuron's output spiketrain of
the form $\frac{\partial S_i[n]}{\partial W_{ij}}$ we
differentiate the network dynamics (Equations~\eqref{eq:syn_discrete_time}
and~\eqref{eq:mem_discrete_time}) and obtain 
\begin{eqnarray}
	\frac{\partial S_k[n+1]}{\partial W_{ij}}&=&
	\Theta^\prime(U_k[n+1]-\vartheta) \left[
		\frac{\partial U_k[n+1]}{\partial W_{ij}} \right] \label{eq:hebb_like_update}\\
	\frac{\partial U_k[n+1]}{\partial W_{ij}}&=& \beta \frac{\partial
	U_k[n]}{\partial W_{ij}} + \frac{\partial I_k[n]}{\partial W_{ij}}
	-\frac{\partial S_k[n]}{\partial W_{ij}} \label{eq:deriv_mem_update}\\
	\frac{\partial I_k[n+1]}{\partial W_{ij}}&=& \alpha \frac{\partial
	I_k[n]}{\partial W_{ij}} + S_j[n] \label{eq:deriv_current_update}
\end{eqnarray}

The above equations define a dynamical system which, given
the starting conditions $S_k[0]=U_k[0]=I_k[0]=0$, can be simulated online and forward in
time to produce all relevant derivatives. 
Crucially, to arrive at useful \glspl{sg}, SuperSpike makes two
approximations. First, $\Theta^\prime$ is replaced by a
smooth surrogate derivative $\sigma^\prime(U[n]-\vartheta)$ (cf.\ Fig.~\ref{fig:surr_partials}). 
Second, the reset term with the negative sign in
Equation~\eqref{eq:deriv_mem_update} is dropped, which empirically leads to
better results.
With these definitions in hand, the final weight updates are given by 
\begin{equation}
\Delta W_{ij}[n] \propto e_i[n] \epsilon \ast \left[ \sigma^\prime(U_k[n]) 
	\frac{\partial U_k[n]}{\partial W_{ij}} \right]
\end{equation}
where $e_i[n] \equiv \epsilon \ast (S_i-S^*_i)$.
These weight updates depend only on  
local quantities \refbox{box:nonlocal}.  

Above, we have considered a simple two-layer network (cf.\
Fig.~\ref{fig:snn_computational_graph}) without recurrent connections. 
If we were to apply the same strategy to compute updates in a \gls{rcnn} or a
network with an additional hidden layer, the equations would become more
complicated and non-local. 
SuperSpike applied to multi-layer networks sidesteps this issue by 
propagating error signals from the output layer
directly to the hidden units as in random \gls{bp} 
(cf.\ Section~\ref{sec:feedback_alignment};
Fig.~\ref{fig:spatial_credit_assignment}c;
\citep{Lillicrap_etal16_randsyna, Baldi_Sadowski16_theoloca,
nokland_direct_2016}).
Thus, SuperSpike achieves temporal credit assignment by propagating all relevant quantities forward in time, while 
it relies on random \gls{bp} to perform spatial credit assignment.

While the work by \citet{zenke_superspike:_2018} was centered around feed-forward networks,
\citet{bellec_biologically_2019} show that similar biologically plausible three factors rule can
also be used to train \gls{rcnn} efficiently.

\subsection{Learning using local errors} 
In practice, the performance of SuperSpike does not scale favorably for large multilayer networks. 
The scalability of SuperSpike can be improved by introducing local errors, as described here.

\begin{figure}[tbhp]
  \centering
  \includegraphics[width=1.0\textwidth]{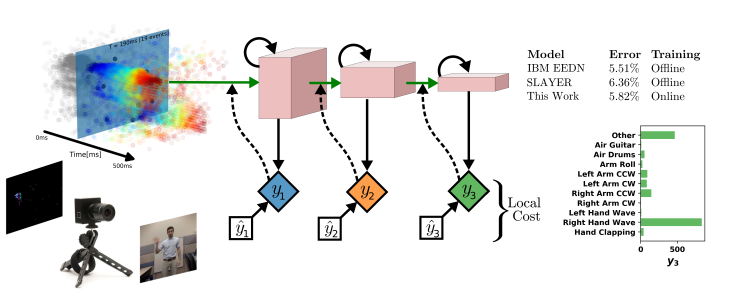}
  \caption{\label{fig:dcll_gestures} Deep Continuous Local Learning (DCLL) with
	spikes \cite{Kaiser_etal18_synaplas}, applied to the event-based DVSGestures dataset. The feed-forward weights (green) of a three layer convolutional \gls{snn} are trained with \gls{sg} using local errors generated using fixed random projections to a local classifier.
	Learning in DCLL scales linearly with the number of neurons thanks to local rate-based cost functions formed by spike-based basis functions.
  The circular arrows indicate recurrence due to the statefulness of the LIF dynamics (no recurrent synaptic connections were used here) and are not trained. This \gls{snn} outperforms BPTT methods \cite{shrestha_slayer:_2018}, requiring fewer training iterations \cite{Kaiser_etal18_synaplas} compared to other approaches.}
\end{figure}

Multi-layer neural networks are hierarchical feature extractors.
Through successive linear projections and point-wise non-linearities, neurons become tuned (respond most strongly) to particular spatio-temporal features in the input.
While the best features are those that take into account the subsequent processing stages and which are learned to minimize the final error (as the
features learned using \gls{bp} do), high-quality features can also be obtained by more local methods.
The non-local component of the weight update equation (Eq.~\eqref{eq:bp_deep})
is the error term $\delta_i^{(l)}[n]$.
Instead of obtaining this error term through \gls{bp}, we require that it be generated using information local to the layer. 
One way of achieving this is to define a layer-wise loss $\mathcal{L}^{(l)}({
	y}^{(l)}[n])$ and use this local loss to obtain the errors.
In such a local learning setting, the local errors $\delta^{(l)}$ becomes:
\begin{align}\label{eq:bp_local}
	\delta_{i}^{(l)} [n] = \sigma'\left(a_i^{(l)}[n] \right)
  \frac{\mathrm{d}}{\mathrm{d}y_i^{(l)}[n]}\mathcal{L}^{(l)}(\mathbf{ y}^{(l)}[n])\text{ where }
	\mathcal{L}^{(l)}(\mathbf{ y}^{(l)}[n]) \equiv \mathcal{L}(\mathbf{ G}^{(l)} \mathbf{
		y}^{(l)}[n],\hat{\mathbf{ y}}^{(l)}[n])
\end{align}
with $\hat{{\bf y}}^{(l)}[n]$ a pseudo-target for layer $l$, and ${\bf G}^{(l)}$ a fixed random matrix that projects the activity vector at layer $l$ to a vector having the same dimension as the pseudo-target.
In essence, this formulation assumes that an auxiliary random layer is attached
to layer $l$ and the goal is to modify $\mathbf{\bf W}^{(l)}$ so as to minimize the discrepancy between the auxiliary random layer's output and the pseudo-target. 
The simplest choice for the pseudo-target is to use the top-layer target. 
This forces each layer to learn a set of features that are able to match the top-layer target after undergoing a fixed random linear projection. Each layer builds on the features learned by the layer below it, and we empirically observe that higher layers are able to learn higher-quality features that allow their random and fixed auxiliary layers to better match the target~\cite{mostafa2018deep}.
\label{sec:spatial_credit_assignment}

A related approach was explored with spiking neural networks \cite{Nicola_Clopath17_supelear}, where separate networks provided high-dimensional temporal signals to improve learning.
Local errors were recently used in \glspl{snn} in combination with the SuperSpike (cf.\ Section~\ref{sec:superspike}) forward method to overcome the temporal credit assignment problem \cite{Kaiser_etal18_synaplas}.
As in SuperSpike, the \gls{snn} model is simplified by using a feedforward structure, and omitting the refractory dynamics in the optimization.
However, the cost function was defined to operate locally on the instantaneous rates of each layer. 
This simplification results in a forward method whose space complexity scales as
$O(N)$ (instead of $O(N^3)$ for the forward method, $O(N^2)$ for SuperSpike, or $O(N T)$ for the backward method), while still making use of spiking neural dynamics. 
Thus the method constitutes a highly efficient synaptic plasticity rule for
multi-layer \glspl{snn}.
Furthermore, the simplifications enable the use of existing automatic differentiation methods in machine learning frameworks to systematically derive synaptic plasticity rules from task-relevant cost functions and neural dynamics (see \cite{Kaiser_etal18_synaplas} and included tutorials), making DCLL easy to implement. 
This approach was benchmarked on the DVS Gestures dataset \reffig{fig:dcll_gestures}, and performs on par with standard \Gls{bp} or \Gls{bptt} rules.

\subsection{Learning using gradients of spike times}
Difficulties in training \glspl{snn} stem from the discrete nature of the quantities of interest such as the number of spikes in a particular interval. The derivatives of these discrete quantities are zero almost everywhere which necessitates the use of \gls{sg} methods. Alternatively, we can choose to use spike-based quantities that have well defined, smooth derivatives. One such quantity is spike times. This capitalizes on the continuous-time nature of SNNs and results in highly sparse network activity as the emission time of even a single spike can encode significant information. Just as importantly, spike times are continuous quantities that can be made to depend smoothly on the neuron's input. Working with spike times is thus a complementary approach to \gls{sg} but which achieves the same goal: obtaining a smooth chain of derivatives between the network's outputs and inputs.
For this example, we use non-leaky integrate and fire neurons described by:

\begin{align}
\label{eq:model_neuron}
	\frac{\mathrm{d}U_i}{\mathrm{d}t} =  I_i \quad \text{with} \quad 
I_i =  \sum\limits_j W_{ij}\sum\limits_r \Theta(t - t_i^r)\exp\left(-(t-t_i^r)\right)
\end{align}
where $t_i^r$ is the time of the $r^\mathrm{th}$ spike from neuron $j$, and $\Theta$ is the Heaviside step function.

\begin{figure}[tbhp]
    \centering
    \subfloat[]{\includegraphics[width=.3\textwidth]{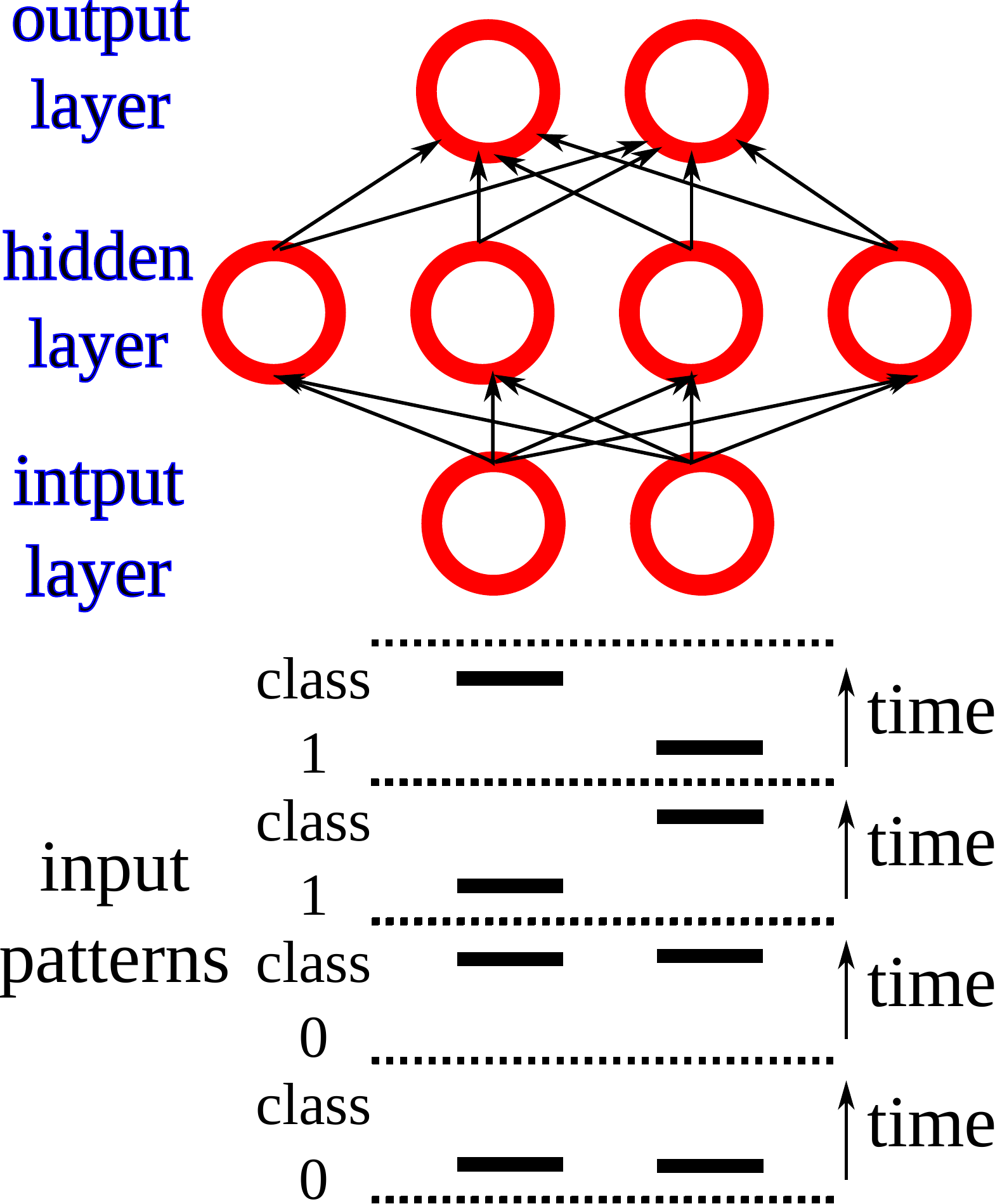} \label{fig:xor_net}}
    \quad
    \subfloat[]{\includegraphics[width=.6\textwidth]{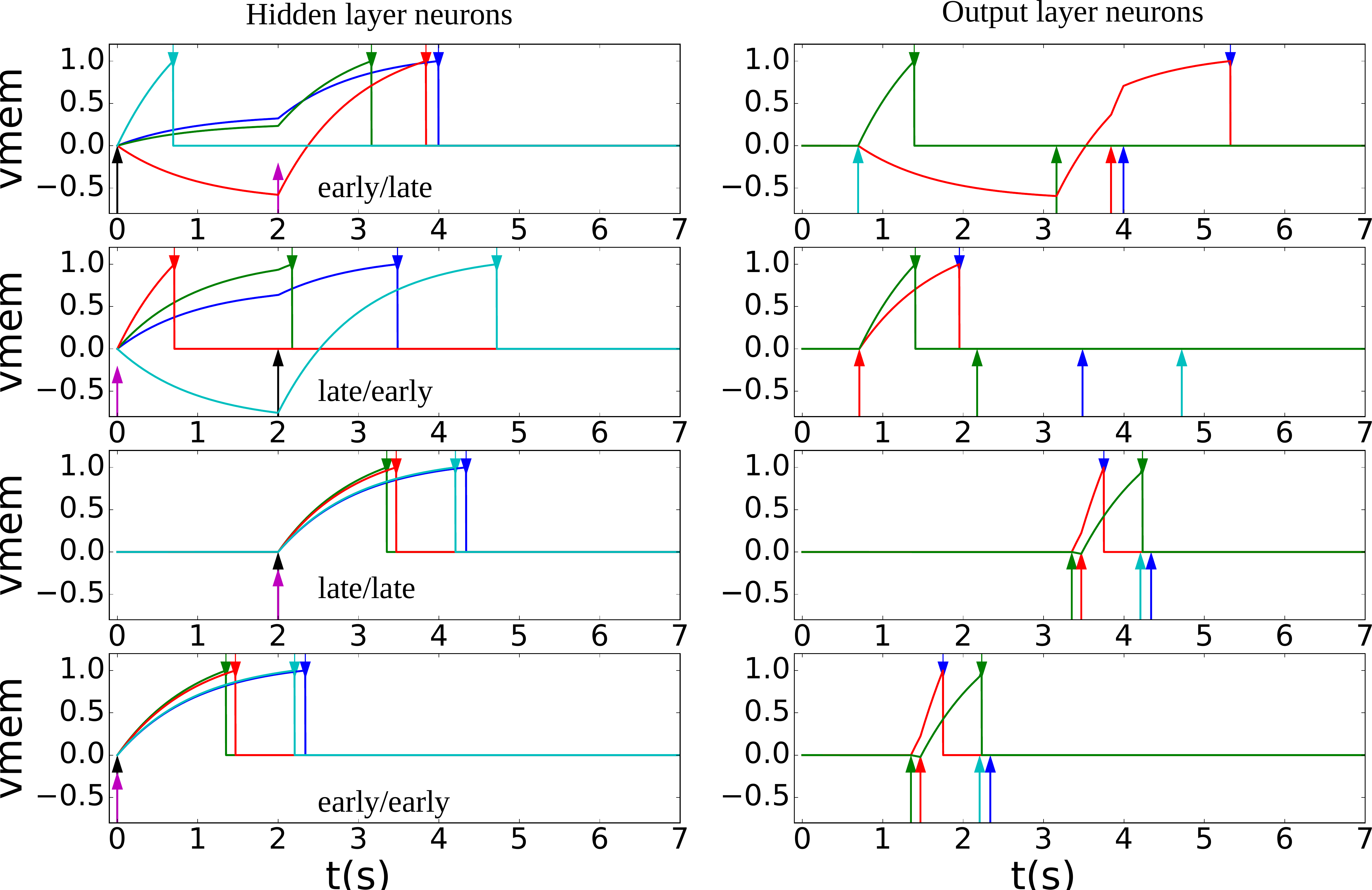} \label{fig:xor_sim}}
    \caption{\textbf{Temporal XOR problem}. (a) An SNN with one hidden layer. Each input neuron emits one spike which can either be late or early resulting in four possible input patterns that should be classified into two classes. (b) For the four input spike patterns (one per row), the right plots show the membrane potentials of the two output neurons, while the left plots show the membrane potentials of the four hidden neurons. Arrows at the top of the plot indicate output spikes from the layer, while arrows at the bottom indicate input spikes. The output spikes of the hidden layer are the input spikes of the output layer. The classification result is encoded in the identity of the output neuron that spikes first. 
}. 
    \label{temporal_xor}
\end{figure}

Consider the simple \textit{exclusive or} (XOR) problem in the temporal domain: A network
receives two spikes, one from each of two different sources. Each spike can
either be ``early'' or ``late''. The network has to learn to distinguish between
the case in which  the spikes are either both early or both late, and the case
where one spike is early and the other is late (Fig.~\ref{fig:xor_net}). 
When designing a SNN, there is 
significant freedom in how the network input and output are encoded. In this
case, we use a first-to-spike code in which we have two output neurons and the
binary classification result is represented by the output neuron that spikes first. 
Figure~\ref{fig:xor_sim} shows the network's response after training (see
\cite{Mostafa16_supelear} for details on the training process). For the first input class (early/late or late/early), one output neuron spikes first and for the other class (early/early or late/late), the other output neuron spikes first. 

\section{Conclusion}
We have outlined how \glspl{snn} can be studied within the framework
of \glspl{rnn} and discussed successful approaches for training them.
We have specifically focused on \gls{sg} approaches for two reasons: \gls{sg} approaches are able to train \glspl{snn} to unprecedented performance levels on a range of
real-world problems. This transition marks the beginning of an exciting time in
which \glspl{snn} will become increasingly interesting for applications which
were previously dominated by \glspl{rnn};  \glspl{sg} provide a framework that ties together ideas from 
machine learning, computational neurosciences, and
neuromorphic computing. 
From the viewpoint of computational neuroscience, the approaches presented in this paper are appealing because several of them are related to ``three-factor'' plasticity rules which are an important class of rules believed to underlie synaptic plasticity in the brain.
Finally, for the neuromorphic community, \gls{sg} methods provide a way to learn under various constraints on communication and storage which makes \gls{sg} methods highly relevant for learning on custom low-power neuromorphic devices. 

The spectacular successes of modern \glspl{ann} were enabled by algorithmic and
hardware advances that made it possible to efficiently train large \glspl{ann}
on vast amounts of data. With temporal coding, \glspl{snn} are universal
function approximators that are potentially far more powerful than \glspl{ann}
with sigmoidal nonlinearities.
Unlike large-scale \glspl{ann}, which had to wait for several decades until the
necessary computational resources were available for training them, we currently
have the necessary resources, whether in the form of mainstream compute devices
such as CPUs or GPUs, or custom neuromorphic devices, to train and deploy
large \glspl{snn}.
The fact that \glspl{snn} are less widely used than \glspl{ann}
is thus primarily due to the algorithmic issue of trainability. In this
article, we have provided an overview of various exciting developments that are
gradually addressing the issues encountered when training \glspl{snn}. Fully
addressing these issues would have immediate and wide-ranging implications, both
technologically, and in relation to learning in biological brains.  

\section*{Acknowledgments}
This work was supported by the Intel Corporation (EN);
the National Science Foundation under grant 1640081 (EN);
the Swiss National Science Foundation Early Postdoc Mobility Grant P2ZHP2\_164960 (HM) ;
the Wellcome Trust [110124/Z/15/Z] (FZ).

\small


\end{document}